\documentclass[10pt,twocolumn,letterpaper]{article}

\usepackage[pagenumbers]{cvpr} % To force page numbers, e.g. for an arXiv version

\usepackage{graphicx}
\usepackage{amsmath}
\usepackage{amssymb}
\usepackage{booktabs}
\usepackage[pagebackref,breaklinks,colorlinks]{hyperref}

\usepackage{pifont} % for cmark and xmark
\newcommand{\cmark}{\ding{51}}%
\newcommand{\xmark}{\ding{55}}%
         
\usepackage{graphicx}
\usepackage{booktabs}
\usepackage{multirow}
\usepackage{makecell}
\usepackage{caption}
\usepackage{threeparttable}
\usepackage{algorithm} 
\usepackage{algorithmic} 
\usepackage{listings}
\usepackage[dvipsnames]{colortbl}
\usepackage[dvipsnames]{xcolor} 
\definecolor{Gray}{gray}{0.85}
\usepackage{hhline}
\def\eg{\textit{e.g. }}
\def\ie{\textit{i.e. }}

\def\etc{\textit{etc. }}

\usepackage[capitalize]{cleveref}
\crefname{section}{Sec.}{Secs.}
\Crefname{section}{Section}{Sections}
\Crefname{table}{Table}{Tables}
\crefname{table}{Tab.}{Tabs.}

\newcommand{\Figref}[1]{{Fig.~\ref{#1}}}

\begin{document}

%%%%%%%%% TITLE - PLEASE UPDATE
\title{ETAD: Training Action Detection End to End on a Laptop}

\author{
Shuming Liu$^{1}$, 
\quad Mengmeng Xu$^{1}$,
\quad Chen Zhao$^{1}$,
\quad Xu Zhao$^{2}$,
\quad Bernard Ghanem$^{1}$
\and
$^{1}$King Abdullah University of Science and Technology 
\quad $^{2}$Shanghai Jiao Tong University\\
{\tt\small \{shuming.liu, mengmeng.xu, chen.zhao, bernard.ghanem\}@kaust.edu.sa}
\quad {\tt\small zhaoxu@sjtu.edu.cn}
}
\maketitle

\begin{abstract} Temporal action detection (TAD) with end-to-end training often suffers from the pain of huge demand for computing resources due to long video duration.
In this work, we propose an efficient temporal action detector (ETAD) that can train directly from video frames with extremely low GPU memory consumption. Our main idea is to minimize and balance the heavy computation among features and gradients in each training iteration. We propose to sequentially forward the snippet frame through the video encoder, and backward only a small necessary portion of gradients to update the encoder. To further alleviate the computational redundancy in training, we propose to dynamically sample only a small subset of proposals during training. Moreover, various sampling strategies and ratios are studied for both the encoder and detector.
ETAD achieves state-of-the-art performance on TAD benchmarks with remarkable efficiency. On ActivityNet-1.3, training ETAD \textbf{in 18 hours} can reach \textbf{38.25\% average mAP} with only \textbf{1.3 GB memory consumption} per video \textbf{under end-to-end training}. Our code will be publicly released.
\end{abstract}

\section{Introduction}
\label{sec:introduction}
Let us assume a junior researcher, who does not have access to a high-end GPU (\eg NVIDIA A100),  starts to research problems in video localization, such as temporal action detection (TAD), which takes a raw video as input and predicts the period of pre-defined temporal activities \cite{caba2015activitynet,Escorcia2016DAPsDA,zhao2019hacs,xu2020g}.
Although great progress has been made in this area, training the whole TAD pipeline is getting computationally heavier and slower, which may discourage the disadvantaged researcher when only limited resources are available. 
To help with this situation, an efficient end-to-end TAD method with a low-cost requirement (\eg a standard laptop) is in demand.

\begin{figure}[t]
    \centering
    \includegraphics[trim={8cm 3.5cm 8.8cm 5cm},width=\linewidth,clip]{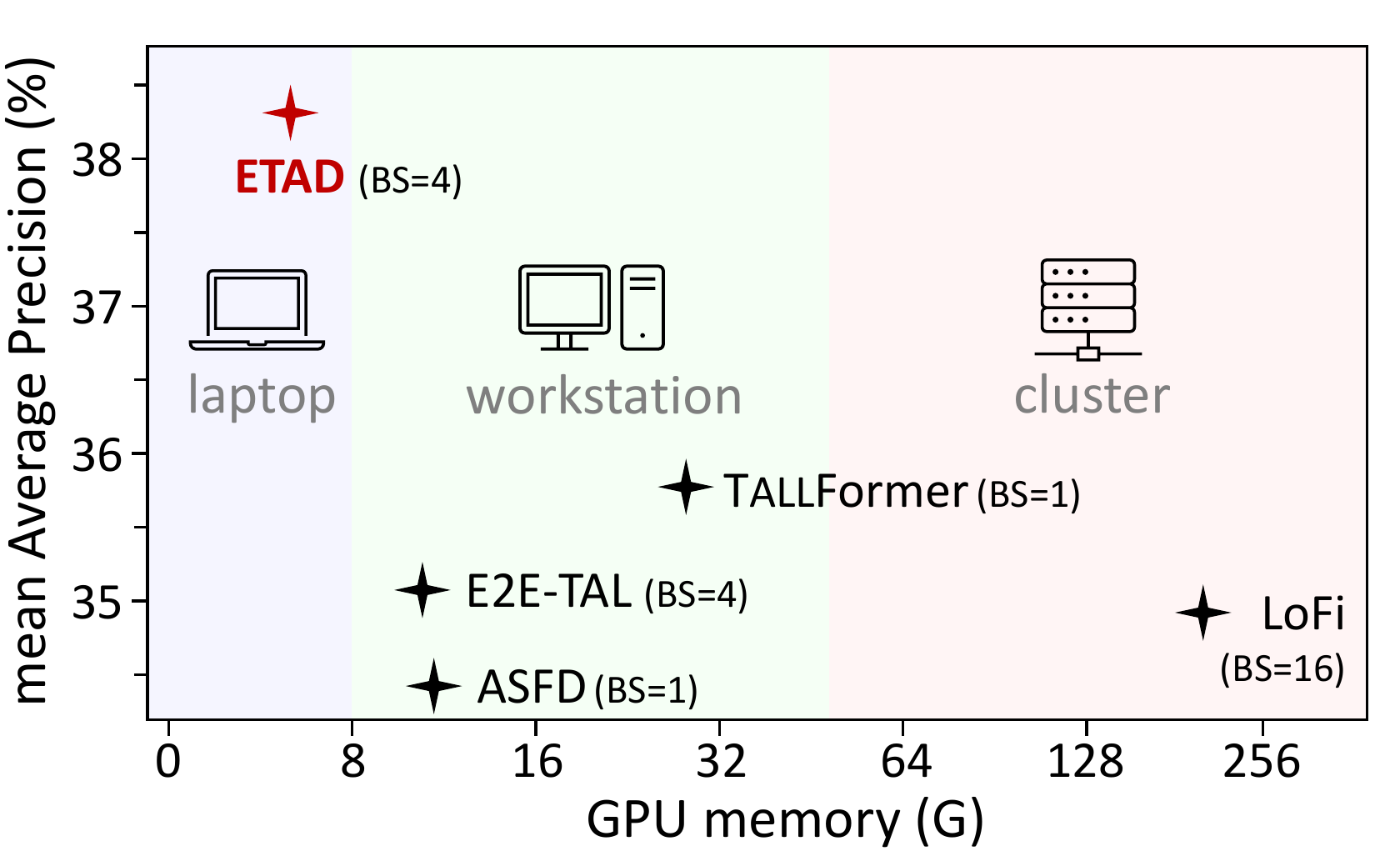}
    % \vspace{-0.7cm}
    \caption{\textbf{Compared with recent end-to-end TAD methods, ETAD has very low GPU memory consumption and SOTA performance.} ETAD minimizes and balances the heavy computation among features and gradients. On ActivityNet-1.3, it reaches {38.25\% average mAP}, $2.65\%$ higher than SOTA end-to-end method TALLFormer~\cite{cheng2022tallformer}, while only using 5.2 GB GPU memory (batch size 4) and 18 hours of training.}
    \label{fig:intro}
\end{figure}

Most of the current TAD pipeline consists of a video encoder and an action detector. Training them jointly, \ie \textit{end-to-end training}, has become the recent trend~\cite{lin2021learning,liu2022empirical,cheng2022stochastic,cheng2022tallformer}. The advantage of such a paradigm is multi-fold, \eg it allows feature adaptation for the target data domain, enables online augmentation to enhance the representation, \etc The main challenge of end-to-end training for TAD is the tremendous GPU memory requirement (\eg 34 GB) to process a single long untrimmed video (\eg 5 mins). This is why TAD methods like ASFD~\cite{Lin_2021_CVPR} resort to downscaling the video frame resolution to $96\times96$ and sampling a small set  of frames (768) during training, while SBP~\cite{cheng2022stochastic} stops a portion of the gradient flow for backpropagation, and TALLFormer~\cite{cheng2022tallformer} caches most of the video features and only updates $15\%-60\%$ of them. Nonetheless, these methods still need moderate GPU memory (\eg 32 GB) to achieve state-of-the-art detection performance.

Our main motivation is to reduce the computation redundancy and leverage minimal GPU memory during end-to-end TAD training. First, although current methods process multiple video snippets in parallel to extract features, our study shows that a sequential process of snippet encoding as well as backpropagation can significantly reduce the peak memory usage without sacrificing any detection performance, and it only moderately increases training time. 
Meanwhile, we observe that not all snippets are needed for updating the video encoder during backpropagation, since most consecutive frames in an untrimmed video are similar in semantics.
Second, to guarantee a high recall rate and cover all potential temporal activities, common TAD practice utilizes a dense distribution of action candidates or proposals, such as the proposal map in BMN \cite{lin2019bmn}. We find that such a design choice is not necessary, since most proposals overlap with each other, and they eventually share similar feature representations.
Our study shows that sampling only a small portion of these proposals does not affect the detection performance but can improve the training efficiency, and further reduce memory usage.

In this work, we propose an Efficient Temporal Action Detector (ETAD), which provides an end-to-end TAD solution that requires extremely low GPU memory and has affordable training time, as shown in Fig.~\ref{fig:intro}. The success of ETAD is based on a \textbf{Sequencialized Gradient Sampling (SGS)} process and an \textbf{Action Proposal Sampling (APS)} design. SGS forwards the snippet frames in micro-batches through the video encoder, and selectively backwards only a small portion of gradients, reducing the peak GPU memory usage by 92\%. Additionally, SGS can reduce the delay in synchronizing the encoder input and 
the detector input, which can result in a training time that is similar to parallelized solutions (only {+14\%}).
APS, on the other hand, generates a much smaller but sufficient set of action candidates during training. It shows that only 6\% of proposals can still guarantee decent action detection performance and greatly remove the training memory redundancy. 
Subsequently, our empirical study on sampling strategies in both modules shows that most common strategies, such as label-guided sampling and feature-guided sampling, are not evidently better than heuristic stochastic sampling, which is the most efficient.

ETAD achieves state-of-the-art performance on two popular benchmarks in an end-to-end fashion with low memory cost and acceptable time consumption. On ActivityNet-1.3, for example, we train ETAD on a single GPU for {18 hours} to reach \textbf{38.25\% average mAP}, $2.65\%$ higher than the end-to-end SOTA method TALLFormer~\cite{cheng2022tallformer}, while the memory consumption is only 1.3 GB per video. Note that this memory usage is even less than many TAD methods that take as input pre-extracted video features \cite{zhao2021video,xu2020g}. 
The main contributions of this work can be summarized as follows:
\begin{enumerate}
\setlength{\itemsep}{0pt}
\setlength{\parsep}{0pt}
\setlength{\parskip}{0pt}
\item We propose to sequentially backpropagate a small portion of gradients to update the video encoder for end-to-end TAD training. This significantly reduces GPU memory usage without increasing training time much.
\item We adopt various sampling strategies to study the snippet gradient redundancy and action proposal redundancy in the current TAD framework. Surprisingly, using only 6\% of proposals and 30\% of snippet gradients can guarantee a good detection performance.
\item Extensive experiments show that ETAD reaches state-of-the-art performance on two TAD benchmarks, ActivityNet-1.3 and THUMOS-14. In particular, ETAD achieves 38.25\% average mAP on ActivityNet with only 1.3 GB per video in end-to-end training.
\end{enumerate}
\section{Related Work}
\label{sec:related_work}
\label{related_tal}

\noindent{\textbf{Temporal Action Detection. }}An action detector can  localize action instances directly from videos (\textit{direct}), or merely refine the boundaries of proposals from a proposal-generation network (\textit{refinement}). The \textit{direct} methods usually focus on enhancing the temporal feature representation \cite{Long2019GaussianTA,xu2020g} or improving the proposal evaluation \cite{heilbron2017scc,xu2017r,chao2018rethinking,lin2019bmn,Bai2020bcgnn}. For example, G-TAD \cite{xu2020g} utilizes graph convolutions to model the correlations between video snippets.
The \textit{refinement} methods tend to prune off-the-shelf action proposals \cite{Escorcia2016DAPsDA,Gao2018CTAPCT,Liu2019MultiGranularityGF} and provide more accurate boundary predictions \cite{Shou2016TemporalAL,qing2021temporal,zhao2021video}.
P-GCN \cite{Zeng2019GraphCN} is a typical refinement method that exploits proposal-proposal relations to refine predictions of BSN \cite{lin2018bsn}. TCANet \cite{qing2021temporal} uses high-quality proposals generated from BMN \cite{lin2019bmn} and proposes a cascade structure to progressively refine actions. Our proposed ETAD belongs to the family of direct solutions, since it  does not rely on any external proposal generation methods, but it surpasses the best refinement method.

\vspace{2pt}\noindent{\textbf{End-to-end Solutions in TAD.}} Recently, more methods study TAD directly from the original video frame to the proposal prediction, which is referred to as \textit{end-to-end training}. The early work R-C3D \cite{xu2017r} encodes the frames with 3D filters and proposes action segments then classifies and refines them. PBRNet \cite{liu2020progressive} and ASFD \cite{lin2021learning} also train detectors from raw frames, but they suffer from the small batch size and low-resolution frames. E2E-TAL \cite{liu2022empirical} further confirms the benefit of end-to-end training for TAD and studies different design choices. 
Moreover, some works propose ways of pre-training the video encoder by new training tasks to close the gap between action recognition and action detection, \eg TSP  \cite{alwassel2020tsp} and BSP \cite{xu2020boundary}. 
Differently, our ETAD is able to train the network with high frame resolution, large batch size, and single-stage training.

\vspace{2pt}\noindent{\textbf{Sampling in Video Understanding.}} Although densely sampling snippets over the entire video is effective for understanding short video clips, such an approach is expensive for long untrimmed videos. 
An alternative way is to summarize the video \cite{huang2019novel} by selecting only the relevant frames or snippets.
% An alternative approach is to select only the relevant frames or snippets, such as in video summarization \cite{huang2019novel}. 
% In action recognition,
For example, SCSampler \cite{korbar2019scsampler} selects salient clips from video for efficient action recognition. SBP \cite{cheng2022stochastic} stochastically drops certain backpropagation paths to train the action recognition/detection model memory efficiently. However, the forward path of SBP still requires a lot of memory for long video input. To reduce the forward computation, TALLFormer \cite{cheng2022tallformer} first stores the pre-computed video feature in a feature bank and only updates a relatively small portion of features in each iteration. Since features in the bank are not always up-to-date, if the training dataset is too large, this method may fail because the features of the same video between two epochs can be drastically different. Besides, proposal sampling in TAD is an under-explored topic, and most methods \cite{lin2019fast,lin2019bmn,xu2020g} exhaustively enumerate the possible locations of activity, leading to redundant computation for highly overlapped proposals.

\section{Method}
\label{sec:method}

Given an untrimmed video, temporal action detection aims to predict its foreground actions, denoted as ${{\Psi }}=\left\{\varphi _i\!=\!(t_s,t_e,c)\right\}_{i=1}^M$, where $({t_s},{t_e},c)$ are the start time, end time, and category of the action instance ${{\varphi }_{i}}$, respectively. $M$ is the total number of actions.

\vspace{4pt}
\subsection{Model Architecture}
\label{sec:architecture}

\begin{figure*}[t]
% \vspace{-3.0em}
\hsize=\textwidth
    \centering
    \includegraphics[trim={2.5cm 6cm 4cm 5cm},width=0.99\textwidth,clip]{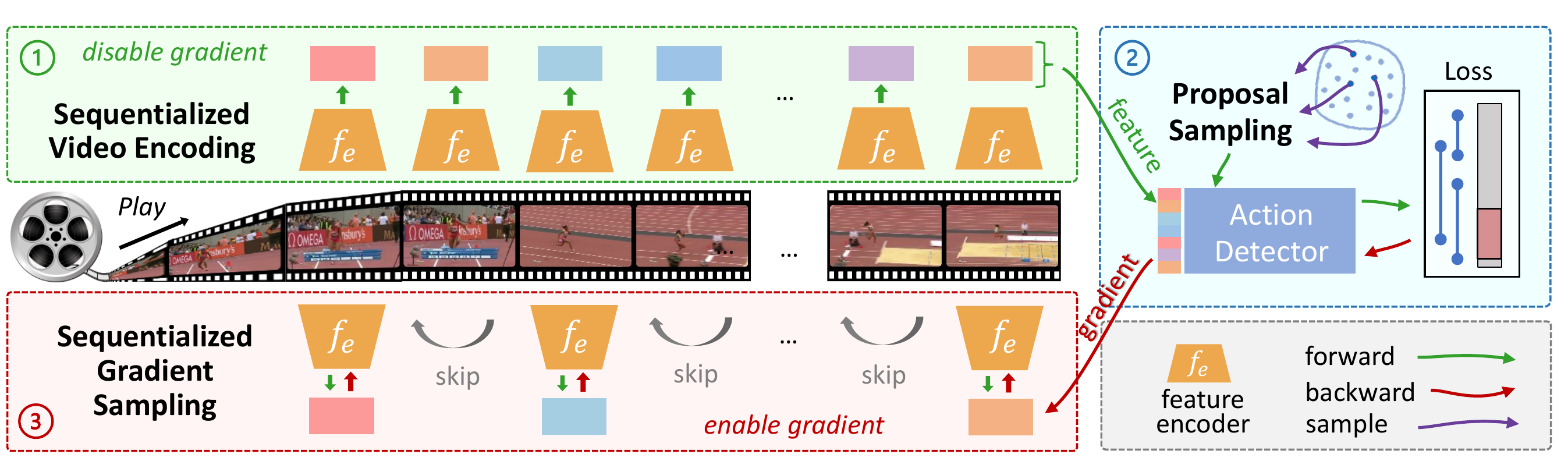}
    % \vspace{-0.5cm}
    \caption{\textbf{The training pipeline of ETAD can be divided into three stages: sequentialized video encoding, action detector learning,  and sequentialized gradient updating.} The sequential video decoding in stage 1 and sequential gradient updating in stage 3 visibly alleviate the whole network's  GPU memory consumption. The action proposal sampling in stage 2 and gradient sampling in stage 3 further cut down the computation redundancy and reduce the training time.} 
    % \vspace{-0.3cm}
\label{fig:method}
\end{figure*}

The overall architecture of ETAD is shown in \Figref{fig:method}, which illustrates the pipeline of feature extraction and action detection. 
For \emph{feature extraction}, an off-the-shelf action recognition model, such as TSM \cite{lin2019tsm}, R(2+1)D \cite{tran2018closer}, is adapted to encode multiple video snippets to a list of feature vectors. Specifically, each vector is obtained from the feature map before the classification head of the recognition model, with global average pooling applied on the temporal and spatial dimensions.
For \emph{action detection}, we adopt a simple yet effective detector to retrieve actions. First, two LSTM layers capture long-range temporal relations to enhance the snippet-level feature representations. Then, two convolution layers are applied to classify the startness and endness of each snippet. Last, a proposal evaluation module refines the candidate proposal boundaries and predicts the proposal confidence. To improve the regression precision of the boundary, we can stack more proposal evaluation modules with progressively improved IoU thresholds. 

While more memory-efficient than existing methods, our action detector with end-to-end training can achieve SOTA results. Based on the simple setup, we adopt sequentialized gradient sampling and action proposal sampling to alleviate the computation burden, targeting efficient end-to-end TAD training with \emph{minimal} memory usage.

\subsection{Preliminary of Sequentialized Gradient}
Typically in TAD, the untrimmed long video is represented as $X\in\mathbb{R}^{N\times 3 \times T \times H \times W}$, where $N$ snippets (or clips) are sampled from the video, and each snippet $x$ has $T$ frames with spatial resolution $H \times W$. Given an arbitrary video encoder denoted as $\boldsymbol{f_{e}}$, which can be a CNN-based or a transformer-based action recognition model, snippet $x$ is encoded as a feature vector $f\in\mathbb{R}^{C}$ by the video encoder. Thus, the feature sequence $F\in\mathbb{R}^{N\times C}$ is extracted from $N$ snippets in parallel. Subsequently, an action detector $\boldsymbol{f_{d}}$ retrieves action candidates $\phi$ from this feature sequence. The whole forward process can be denoted as follows:
\begin{equation}
    F = [f_1,f_2,...,f_N] = \boldsymbol{f_{e}}([x_1,x_2,...,x_N])
\label{eq:encoder}
\end{equation}
\begin{equation}
\phi = \boldsymbol{f_{d}}(F)
\label{eq:detector}
\end{equation}

In order to update the parameters $\boldsymbol\theta$  of video encoder $\boldsymbol{f_{e}}$ during training, all the intermediate activations in Eq.~(\ref{eq:encoder}) must be saved for later gradient backpropagation. Given the loss $L$, the gradient of $\boldsymbol\theta$ are computed  by:
\begin{equation}
\Delta \boldsymbol\theta = \frac{\partial L}{\partial \boldsymbol\theta} = \frac{\partial L}{\partial F} \cdot \frac{\partial F}{\partial \boldsymbol\theta} \\
\label{eq:backward}
\end{equation}

Compared with standard action recognition, the existence of snippet dimension in $X$ causes a tremendous computation in $\boldsymbol{f_{e}}$, which increases linearly with the number of snippets $N$.
For example, when using 128 snippets, $224\times224$ resolution, 16 batch size (\eg 16 videos), even an efficient TSM~\cite{lin2019tsm} model takes more than $500$ GB memory, which is infeasible to perform end-to-end training on most platforms. 
However, the computation graph of each snippet in $\boldsymbol{f_{e}}$  is essentially independent of others. Precisely, as long as no batch-level parameters need to update, such as batch normalization, we can apply 
associative law to Eq.~(\ref{eq:encoder}) to get $f_i=\boldsymbol{f_{e}}(x_i)$,
then the parallel computation in Eq.~(\ref{eq:backward}) can be eventually decoupled as:
\begin{equation}
\Delta \boldsymbol\theta =  \frac{\partial L}{\partial [f_1,...,f_N] } \frac{\partial [f_1,...,f_N] }{\partial \boldsymbol\theta} = \sum_{i=1}^{N}\Delta f_i\frac{\partial f_i }{\partial \boldsymbol\theta}
\label{eq:backward_decouple}
\end{equation}

\subsection{Sequentialized Gradient Sampling}
\label{sgs}
Eq.~(\ref{eq:backward_decouple}) suggests that the computation of the partial derivative of snippet feature $F$ to encoder parameter $\boldsymbol\theta$ can be obtained from two different ways, \ie parallelly compute the derivative from the $N$ snippets in one-step, or divide $N$ snippets into multiple \textbf{micro-batches} (a micro-batch has $K$ snippets, $K<<N$), and then \textbf{\textit{sequentially}} compute the gradient within a micro-batch by $N/K$ iterations and accumulate the gradients. Based on the above derivation, we propose \textbf{Sequentialized Gradient Sampling (SGS)} for efficient end-to-end TAD training, which is consist of three stages as illustrated in Fig. \ref{fig:method}.
\begin{enumerate}
\setlength{\itemsep}{0pt}
\setlength{\parsep}{0pt}
\setlength{\parskip}{0pt}
\item \textit{Sequentialized video encoding.} We temporally split video $X$ into micro-batches. Each micro-batch has $K$ snippets. We run forward passes on the encoder in eval mode for $N/K$ times, and concatenate the output features as $F$.
\item \textit{Action detector learning.} We use $F$ to train the detector for one iteration and backpropagate the gradients to the concatenated features $F$. We collect the feature gradients $\Delta [f_1,\cdots,f_N]$, and free all the cache in GPU memory.
\item \textit{Gradient sampling and sequentialized updating.} We sample $\gamma$ portion of the feature gradients to train the encoder in a sequential fashion. In each step, we use a micro-batch to compute the gradients of encoder parameters and accumulate all the gradients for the later parameter update. 
\end{enumerate}

The key to achieving efficient end-to-end training by SGS is to sequentially process a small micro-batch data in each iteration during stage 1 and stage 3, and only backward a small portion of gradients during stage 3. As a comparison, in traditional end-to-end training,
all $N$ snippet intermediate activations in the video encoder are reserved for later backpropagation, which takes over 95\% of total GPU memory. Instead, our SGS operates on a small data volume for video encoding during each iteration, and the peak memory usage is only $\frac{K}{N}$ of the traditional end-to-end setting. Since the micro-batch data is related to $K$ instead of $N$, such memory usage can be constant and independent of the video length. In the extreme case $K\!=\!1$, no matter how long the video is given, the maximum memory usage by SGS is aligned with the memory usage as designed in the action recognition task. 
Another advantage of SGS is the high GPU utilization, since it does not require all the $N$ snippets to be ready (\eg video decoding, data augmentation, \etc), which may cause a high latency in the traditional parallel design.

Moreover, although extra forward computation is involved in stage 3, using gradient sampling in SGS can address this deficiency and reduce the overall computation to be less than the original end-to-end training (see Tab. \ref{tab:comparison_e2e}). In the experiments, we find that such sampling won't affect the TAD performance (see Sect. \ref{ablation}). This is because the consecutive video frames in the untrimmed video are usually similar in appearance and semantics, and the feature vectors of corresponding snippets may share similar representations. Thus, the gradients of encoders on such snippets are similar. Moreover, as mentioned by \cite{cheng2022tallformer}, since the video encoder is already pre-trained on a large-scale action recognition dataset, thus it evolves more slowly than other modules
in the network with a smaller learning rate, leading to relatively small gradient values. Based on such insight, our SGS which only backpropagates a small ratio of snippets would still guarantee high TAD performance.

Regards to time efficiency, although the sequential processing breaks down the parallel design, the total training time using SGS is only 114\% than original end-to-end training, but it requires less than 1/25 memory of the default setting (see Tab.~\ref{tab:training_time}). Besides, SGS can be complementary of other memory-efficient techniques, such as activation checkpointing \cite{chen2016training}, mixed-precision training \cite{micikevicius2017mixed}, \etc Noted that our SGS is agnostic of encoder architecture and thus can incorporate any of the common encoders in its framework. The pseudo-code of our SGS algorithm can be found in the \textit{supplementary material}.

\subsection{Action Proposal Sampling}

Beyond SGS,  we also study proposal sampling, which aims to reduce the redundant action proposals in the action detector. In the current two-stage TAD methods (\ie methods use RoI alignment or similar to extract proposal features explicitly), a dense candidate proposal set is needed in the second stage for proposal refinement and post-processing. For example, BMN \cite{lin2019bmn} and G-TAD \cite{xu2020g} propose to enumerate all possible combinations of start and end locations as candidate proposals to deal with the large action length variation. Mathematically, given the number of snippets $T$, there will be $C_{T}^{2} = T \cdot (T-1) / 2 $ proposals, which has the quadratic complexity with respective to $T$.
However, due to the dense enumeration, most of these proposals overlap with each other. Thus, a large portion of the extracted proposal features is similar or duplicated. Moreover, the proposal evaluation module in TAD usually refines each proposal's start and end boundary~\cite{liu2020progressive}, so it is unnecessary to consider proposals that are temporally close. 
To reduce such redundancy while preserving performance, we propose to replace the densely sampled proposal set  with a subset produced by an efficient sampling, called \textbf{Action Proposal Sampling (APS)}, as illustrated in \Figref{fig:method}. 
Our experiments suggest that with a proper sampling strategy (see Sect.~\ref{sec:howto_sample}), using only 6\% proposals can provide a similar detection performance to the full setup, but it saves more than 90\% of the detector's computation. 

We show that combining APS and SGS can be extremely memory-efficient when training the end-to-end action detector, in the meantime, we can achieve state-of-the-art detection performance.

\subsection{Sampling Strategy}
\label{sec:howto_sample}

We further study the possible sampling strategies in both SGS and APS. Three types of sampling strategies are proposed and compared: \textit{heuristic sampling}, \textit{feature-guided sampling}, and \textit{label-guided sampling}.

\begin{figure}[t] 
\begin{center}
\includegraphics[width=\linewidth]{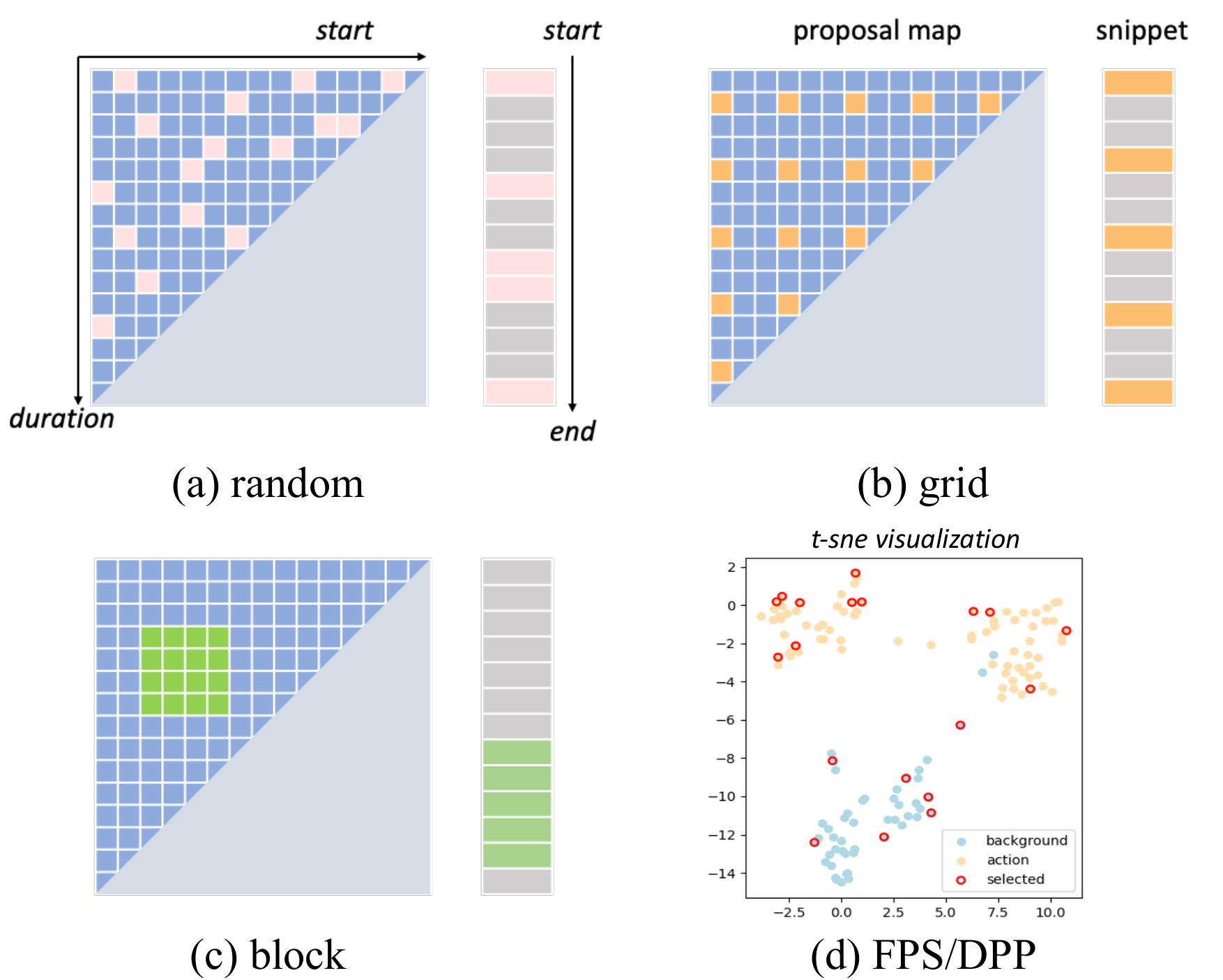}
\end{center}
% \vspace{-0.4cm}
\caption{\textbf{Different sampling strategies: \textit{random}, \textit{grid}, \textit{block}, \textit{FPS}, \textit{DPP}.} \textit{Random} strategy samples certain proposals/snippets from a uniform distribution, \textit{grid} strategy samples with a fixed temporal stride, and \textit{block} strategy samples a consecutive area in proposal map/snippet sequence. Besides, \textit{FPS} (farthest point sampling) chooses the new snippet/proposal which has the farthest distance to the selected samples. \textit{DPP} (determinantal point process) selects the data from the feature embedding space to enforce the sample diversity.}
\label{fig:sampling_strategy}
% \vspace{-0.5cm}
\end{figure}

\noindent{\textbf{Heuristic Sampling}} includes three strategies: random, grid, and block. They are similar to the samplings in MAE~\cite{he2022masked}.
The \textit{random} strategy simply samples snippets or proposals randomly following a uniform distribution. The \textit{grid} strategy samples the snippets with a pre-defined temporal stride (along the temporal dimension) or grid (on a proposal map), as shown in \Figref{fig:sampling_strategy}.
The \textit{block} strategy samples consecutive snippets or a block of proposals in the proposal map. This strategy essentially evaluates the model in a trimmed clip of the video. 

\noindent{\textbf{Feature-guided Sampling}} are based on the data distribution in the feature space. Farthest Point Sampling (FPS, \cite{eldar1997farthest}) selects the new snippet/proposal which has the farthest distance, where a distance is defined as the euclidean distance between two snippet features or proposal features. 
FPS can provide the most distinguished samples of the candidates since the selected samples are more variant in the embedding space.
We also implement the Determinantal Point Process (DPP) to enforce diversity during training. We take the cosine similarity as the kernel function and update the determinant in every training epoch. When a sampling ratio is given, we can directly apply kDPP \cite{kulesza2011k} because the target has a fixed size. Please refer to the \textit{supplementary material} for implementation details.

\noindent{\textbf{Label-guided Sampling}} uses ground truth supervision during action proposal sampling. IoU-balanced sampling \cite{pang2019libra} guarantees the selected proposals have nearly the same number in different IoU thresholds, such as $\{0,0.3,0.7,1\}$. Similarly, scale-balanced sampling maintains the equivalence of proposal numbers around different action scales: small (scale$<$0.3), middle (0.3$<$scale$<$0.7), and large (scale$>$0.7).

Our experiment shows that random sampling, grid sampling, and DPP-based sampling all work well in performance. Using a rather small sampling rate, \eg 30\% at SGS and 6\% at the APS, can provide a decent TAD performance. Such small sampling ratios can greatly reduce computation in both the video encoder and the action detector while keeping the SOTA performance.

\begin{table*}[th]
\centering
\caption{\textbf{Action localization results on the validation set of ActivityNet-1.3}, measured by mAP (\%) at different tIoU thresholds and the average mAP. E2E means the method is under end-to-end training. Mem. is the GPU memory usage  \textit{per video}.}
\small
\begin{tabular}{lccc|ccc>{\columncolor[gray]{0.9}}cc|c}
\toprule
Method                                        & Video Encoder & E2E & Flow &  0.5  &  0.75  & 0.95 & Average & Mem. (GB) & Pub. \\
\midrule
RTD-Action~\cite{tan2021relaxed}              & I3D                  & \xmark  & \cmark  & 47.21 & 30.68 & 8.61 & 30.83  & - & ICCV2021 \\
P-GCN~\cite{Zeng2019GraphCN}                  & I3D                  & \xmark  & \cmark  & 48.26 & 33.16 & 3.27 & 31.11  & - & ICCV2019 \\
BMN~\cite{lin2019bmn}                         & TSN                  & \xmark  & \cmark  & 50.07 & 34.78 & 8.29 & 33.85  & - & ICCV2019 \\
VSGN~\cite{zhao2021video}                     & TSN                  & \xmark  & \cmark  & 52.38 & 36.01 & 8.37 & 35.07  & 1.6 & ICCV2021 \\
G-TAD~\cite{xu2020g}                          & R(2+1)D-34 (TSP)     & \xmark  & \xmark  & 51.26 & 37.12 & 9.29  & 35.81 & 0.7 & CVPR2020 \\
CSA~\cite{sridhar2021class}                 & R(2+1)D-34 (TSP)     & \xmark  & \xmark  & 52.64 & 37.75 & 7.94  & 36.25 & - & ICCV2021 \\
ActionFormer~\cite{zhang2022actionformer}     & R(2+1)D-34 (TSP)     & \xmark  & \xmark  & 54.70 & 37.80 & 8.40  & 36.60 & - & ECCV2022 \\
RCL~\cite{wang2022rcl}                        & R(2+1)D-34 (TSP)     & \xmark  & \xmark  & 55.15 & 39.02 & 8.27  & 37.65 & - & CVPR2022 \\
\midrule
% SSN~\cite{zhao2017temporal}  & ICCV2017 \\
R-C3D~\cite{xu2017r} & C3D &\cmark &\xmark & - & - & - & 26.80 & - & ICCV2017 \\
AFSD~\cite{lin2021learning}                                          & I3D                  & \cmark  & \cmark   & 52.40 & 35.30 & 6.50 & 34.40 & 12 & CVPR2021\\
LoFi~\cite{xu2021low}                        & TSM-ResNet50         & \cmark  & \xmark   & 50.91 & 35.86 & 8.79 & 34.96 & 29 & NeurIPS2021 \\
PBRNet~\cite{liu2020progressive}             & I3D                  & \cmark  & \cmark   & 53.96 & 34.97 & 8.98 & 35.01 & - & AAAI2020 \\
E2E-TAL~\cite{liu2022empirical}   & SlowFast-ResNet50 & \cmark & \xmark & 50.47 & 35.99 & \textbf{10.83} & 35.10 & 3 & CVPR2022 \\
TALLFormer~\cite{cheng2022tallformer}         & Video Swin-B         & \cmark  & \xmark   & 54.10 & 36.20 & 7.90 & 35.60 & 29 & ECCV2022 \\
% \midrule
\textbf{ETAD}             & TSM-ResNet50         & \cmark  & \xmark   & 53.79 & 37.59 & 10.56 & 36.79 & 1.7 & -\\
\textbf{ETAD}             & R(2+1)D-34 (TSP)     & \cmark  & \xmark   & \textbf{55.49} & \textbf{39.32} & 10.57 & \textbf{38.25} & \textbf{1.3} & - \\
\bottomrule
\label{tab:sota_anet}
\end{tabular}
% \vspace{-8pt}
\end{table*}
\section{Experiments}
\label{sec:experiments}

\subsection{Implementation details}

\noindent{\textbf{Datasets and evaluation metrics.}} ActivityNet-1.3 \cite{caba2015activitynet} is a large-scale video understanding dataset, consisting of 19,994 videos annotated for the temporal action detection task. The dataset is divided into train, validation, and test sets with a ratio of 2:1:1. THUMOS-14 \cite{jiang2014thumos} contains 200 annotated untrimmed videos in the validation set and 213 videos in the test set. We also evaluate our methods on the HACS dataset \cite{zhao2019hacs} and achieve state-of-the-art performance (see \textit{supplementary material}). Mean Average Precision (mAP) at certain IoU thresholds and average mAP are reported as the main evaluation metrics. On ActivityNet-1.3,  the IoU thresholds are chosen from 0.5 to 0.95 with 10 steps. On THUMOS-14, the thresholds are chosen from $\{0.3, 0.4, 0.5, 0.6, 0.7\}$. 
% Performing action detection on this dataset is challenging because over 70\% of the untrimmed video is background and each video has more than 15 action instances on average.

\vspace{4pt}\noindent{\textbf{Implementation Details.}} Our method is implemented with PyTorch 1.12, CUDA 11.1, and mmaction2 \cite{2020mmaction2} on 1 Tesla V100 GPU by default.
TSM \cite{lin2019tsm} and R(2+1)D \cite{tran2018closer} are adopted as our video encoder for end-to-end training on ActivityNet-1.3, while two stream I3D \cite{carreira2017quo} is adopted as the encoder on THUMOS-14. We fix the weights of the first two stages of the video encoder and freeze all batch normalization layers. For TSM, the image resolution is set to $224\times224$ with clip length 8, which is the same as in \cite{xu2021low}. For R(2+1)D, the image resolution is set to $112\times112$, and the clip length is set to 16, following \cite{alwassel2020tsp}. We adopt random cropping as data augmentation. Note that the TSM model is only pretrained on Kinetics-400 \cite{kay2017kinetics} and not finetuned on the target datasets, \ie ActivityNet-1.3, or THUMOS-14. The R(2+1)D model is pretrained on the ActivityNet dataset by \cite{alwassel2020tsp}. We use a batch size of 4 and the AdamW optimizer \cite{loshchilov2017decoupled} with weight decay of $10^{-4}$. The learning rate is set to $10^{-3}$ for the action detector and $10^{-6}$/$10^{-7}$ for TSM/R(2+1)D. The micro-batch size $K$ in SGS is set to 4 by default. The sampling ratios are 30\% and 6\% in SGS and APS, respectively. The total training epoch is set to 6 and the learning rate decays by 0.1 after 5 epochs. Following \cite{lin2019bmn,qing2021temporal}, we apply the video-level classification scores from \cite{zhao2017cuhk} on ActivityNet-1.3 and \cite{wang2017untrimmednets} on THUMOS-14.

\subsection{Comparison with State-of-the-Art Methods}

\noindent\textbf{ActivityNet-1.3.}
Tab.~\ref{tab:sota_anet} compares ETAD with other state-of-the-art methods on ActivityNet-1.3. Under end-to-end training, ETAD achieves 38.25\% average mAP with only 1.3 GB memory (per video), outperforming other state-of-the-art end-to-end training methods both on efficiency and efficacy by a large margin. Compared with LoFi \cite{xu2021low} which also uses TSM-ResNet50 as the video encoder, ETAD achieves +1.83 average mAP gain with only 15\% GPU budget. Interestingly, the memory usage of end-to-end-based ETAD is even smaller than feature-based VSGN~\cite{zhao2021video}, suggesting that ETAD is extremely memory-efficient. When the batch size is 4, ETAD's total memory usage is still lower than 8 GB, which can be easily trained on a RTX2080. 

\vspace{4pt}\noindent\textbf{THUMOS-14.} We also show the advantage of our method on THUMOS-14 in Tab.~\ref{tab:sota_thu}, which reaches the comparable performance with other end-to-end methods, such as ASFD~\cite{lin2021learning}, E2E-TAL~\cite{liu2022empirical}. Particularly, ETAD achieves stronger performance on high IoU thresholds, indicating the high precision of the generated action boundaries. 
Furthermore, our SGS can enable end-to-end training of SOTA feature-based TAD methods, \eg ActionFormer\cite{zhang2022actionformer}. As shown in Tab.~\ref{tab:sota_thu} (bottom block), end-to-end training consistently boosts the mAP under all IoU thresholds, while only costing 10.4 GB memory with the heavy Swin transformer backbone.

\begin{table}[bt]
\caption{\textbf{Action localization results on test set of THUMOS14}, measured by mAP (\%) at different tIoU thresholds. $\dagger$ means the reproduced results with Video Swin-T and only RGB modality.}
\small
\begin{tabular}{l|cc>{\columncolor[gray]{0.9}}ccc}
	\toprule
	Method          & 0.3  & 0.4  & 0.5  & 0.6  & 0.7 \\
	\midrule
	SSN~\cite{zhao2017temporal}   & 51.9 & 41.0 & 29.8 & - & - \\
    BMN~\cite{lin2019bmn}                  & 56.0 & 47.4 & 38.8 & 29.7 & 20.5    \\
    G-TAD~\cite{xu2020g}                  & 57.3 & 51.3 & 43.0 & 32.6 & 22.8    \\
    TCANet~\cite{qing2021temporal}               & 60.6 & 53.2 & 44.6 & 36.8 & 26.7    \\
    % BCNet                & 66.5 & 60.0 & 51.6 & 41.0 & 29.2    \\
    VSGN~\cite{zhao2021video}                 & 66.7 & 60.4 & 52.4 & 41.0 & 30.4    \\
    AFSD~\cite{lin2021learning}               & 67.3 & 62.4 & 55.5 & 43.7 & 31.1    \\
    % MUSES                & 68.9 & 64.0 & 56.9 & 46.3 & 31.0    \\
    % TAGS                 & 68.6 & 63.8 & 57.0 & 46.3 & 31.8 \\
    E2E-TAL~\cite{liu2022empirical}                      & 69.4 & 64.3 & 56.0 & 46.4 & 34.9 \\
    \textbf{ETAD}                & \textbf{69.63} & \textbf{64.47} & \textbf{56.17} & \textbf{47.18} & \textbf{35.89} \\
    \midrule
    {ActionFormer$^\dagger$}                & 69.63 & 62.63 & 51.26 & 38.29 & 21.10 \\
    \textbf{$\cdots+$ETAD}                & \textbf{72.82} & \textbf{66.95} & \textbf{57.28} & \textbf{44.51} & \textbf{28.75} \\
    \bottomrule
\end{tabular}
\label{tab:sota_thu}
% \vspace{-8pt}
\end{table}

\subsection{Ablation Study}
\label{ablation}
In this section, we conduct ablation studies on ActivityNet-1.3 to verify the effectiveness of each design in ETAD.

\begin{figure}[h] 
\begin{center}
\includegraphics[trim={0.5cm 0.5cm 0.5cm 0.5cm},width=8cm,clip]{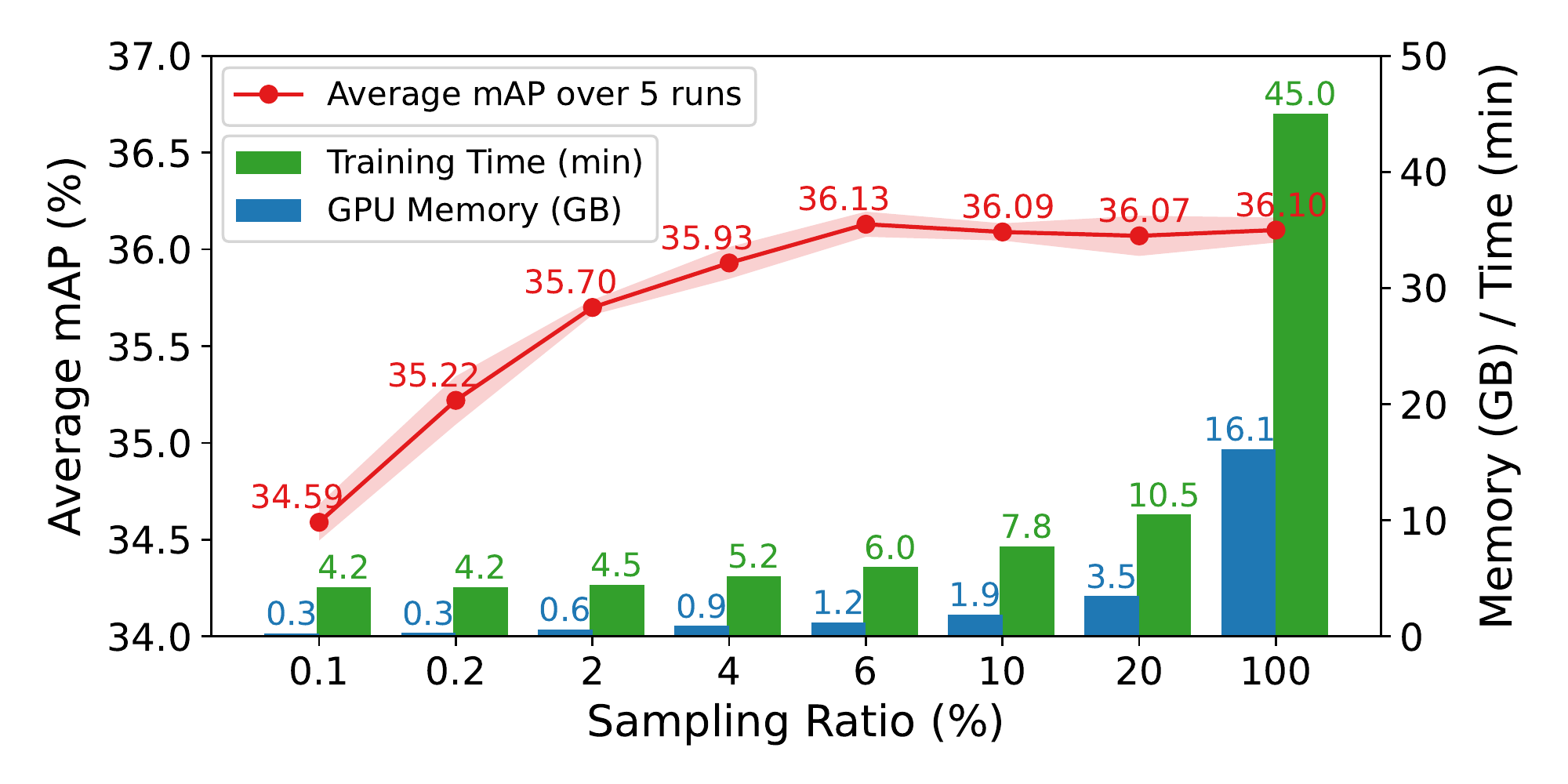}
\end{center}
% \vspace{-5pt}
\caption{\textbf{Using only 6\% proposals are sufficient for action detection.} For proposal sampling, we use pre-extracted TSM features with different sampling ratios and report the mAP, GPU memory, and training time. Random sampling is adopted.}
\label{fig:aps_ratio}
% \vspace{-0.5cm}
\end{figure}

\vspace{4pt}\noindent\textbf{Up to 94\% dense action proposals are redundant for action detection.} To prepare an efficient and powerful action detector for end-to-end training, we first operate on pre-extracted video features to verify the effectiveness of APS. \Figref{fig:aps_ratio} shows that the performance of the detector saturates from a small proposal sampling ratio. When the sampling ratio is under 4\%, the mAP starts to drop visibly. This result confirms our assumption that dense enumerated proposals are redundant for action detection. Using 6\% sampling, ETAD successfully results in the same detection performance as using a complete proposal set. With pre-extracted frozen features, it speeds up the training 7.5x faster (from 45 mins to 6 mins), and cuts down 92\% of memory usage (from 16 GB to 1.2 GB). 

\noindent\textbf{End-to-end training can improve TAD performance, but it is memory-consuming.} Based on the efficient action detector with APS, we extract the snippet feature with a learnable video encoder and jointly optimize it with the action detector. As shown in Tab. \ref{tab:training_time} (first row), end-to-end training can bring a significant performance gain from 36.13 to 36.85, which also proves the importance of end-to-end TAD training. However, this naive end-to-end training requires 137 GB memory, which is infeasible on most platforms. 

\vspace{4pt}\noindent\textbf{Sequential backpropagation is memory-efficient for end-to-end training, but it is also time-consuming.} To further alleviate the memory limitations of end-to-end training, we apply sequential backpropagation on the naive end-to-end training, as shown in Tab.~\ref{tab:training_time} (middle block). In experiments, we also find such an implementation has the same detection performance as the naive one, which also verifies the equality discussed in Sect. \ref{sgs}. Thus, we only compare the peak GPU memory and training time, which shows that adopting sequential backpropagation in end-to-end training can greatly reduce the GPU memory consumption from 103 GB to 3.8 GB. Unfortunately, the training time is also increased by $2.6\times$ larger. Besides, since we need to recompute the activations during backpropagation, the number of FLOPs is also increased.

\begin{table}[t]
\caption{\textbf{Sequential backpropagation can greatly reduce the peak GPU memory while requiring more training time. Combined with gradient sampling, ETAD can achieve efficient and effective end-to-end training.} Seq. means adopting sequentialized backpropagation. Ratio is the gradient sampling ratio. $K$ stands for the micro-batch size in SGS.}
\centering
\small
\setlength{\tabcolsep}{5pt}
\begin{tabular}{ccc|ccc|c}
	\toprule
	Seq.       & Ratio & $K$ & FLOPs & Mem.(GB) & Time & mAP \\
	\midrule
    \xmark & 100\% & - & 100\% & 137 & 100\% & 36.85 \\
	\midrule
    \multirow{4}{*}{\cmark} & \multirow{4}{*}{100\%} & 8 & 150\% & 10.3 & 180\% & \multirow{4}{*}{36.85} \\
    & & 4 & 150\% & 6.6 & 190\%\\
    & & 2 & 150\% & 4.7 & 194\% \\
    & & 1 & 150\% & 3.8 & 264\% \\
    \midrule
    \multirow{5}{*}{\cmark} & 50\% & 4 & 100\% & 6.6 & 137\% & 36.83\\
    & 40\% & 4 & 90\% & 6.6 & 121\% &36.82\\
    & \textbf{30\%} & \textbf{4} & \textbf{80\%} & \textbf{6.6} & \textbf{114\%} & \textbf{36.79} \\
    & 20\% & 4 & 70\% & 6.6 & 101\%  & 36.75 \\
    & 10\% & 4 & 60\% & 6.6 & 91\% & 36.68 \\
    \bottomrule
\end{tabular}
\label{tab:training_time}
\end{table}

\vspace{4pt}\noindent\textbf{Gradient sampling can effectively save training time, without sacrificing detection performance.} To further reduce the training time, gradient sampling is combined with sequential backpropagation, known as our complete SGS approach. As shown in Tab.~\ref{tab:training_time} (bottom), gradient sampling with a ratio larger than 30\% can still maintain nearly the same detection performance, which proves the existence of snippet-level learning redundancy. Such a scenario also happens in THUMOS-14 dataset (see \textit{supplementary material}). In the meantime, the training time is evidently decreased from 190\% to 114\%, which is almost the same as the naive end-to-end training. These results verify that SGS can be served as an effective tool for memory-efficient end-to-end TAD training.

\noindent {\bf{Heuristic sampling strategy is recommended}}. We further study different sampling strategies on ActivityNet-1.3, as shown in Tab.~\ref{tab:strategy}. From the APS column for proposal  sampling, we find that random sampling, grid sampling, and DPP work well. While block sampling and label-guided sampling both show certain downgrades in performance because they change the proposal distribution and thus can not guarantee the variety of proposals. From the SGS column for gradient sampling, all experiments outperform the pre-extracted feature baseline in APS (36.13\%). 
Considering the detection performance and computation complexity of different sampling strategies, we recommend adopting heuristic samplings such as random or grid sampling strategies in APS and SGS. More discussions can be found in the \textit{supplementary material}.

\begin{table}[t]
\centering
% \vspace{-0.1cm}
\captionof{table}{\textbf{Effect of different sampling strategies.} We apply the TSM-R50 as the video encoder and report the mAP on ActivityNet. Frozen backbone is used in APS and end-to-end training is used in SGS, thus the later results are expected to be higher.}
\small
\begin{tabular}{p{1.6cm}<{\centering}p{2.3cm}<{\centering}|p{1.2cm}<{\centering}|p{1.2cm}<{\centering}}
\toprule
\makecell{Sampling\\Type} & \makecell{Sampling\\Strategy} & \makecell{APS} & \makecell{SGS} \\
\midrule   
\multirow{3}{*}{\textit{heuristic}}  
& random      & \textbf{36.13} & \textbf{36.79} \\
& grid        & 36.04 & 36.77 \\ 
& block       & 32.97 & 36.74 \\ 
\midrule   
\multirow{2}{*}{\textit{feature-guided}} 
& FPS & 33.59 & 36.61 \\
& DPP & \textbf{36.16} & \textbf{36.78} \\ 
\midrule   
\multirow{2}{*}{\textit{label-guided}} 
& IoU-balanced   & 34.84 & N.A. \\
& Scale-balanced & 35.10 & N.A. \\
\bottomrule
\end{tabular}
\label{tab:strategy}
% \vspace{-0.2cm}
\end{table}

\subsection{Further Discussions}
\noindent\textbf{Compared with other end-to-end strategies, SGS shows both memory-efficiency and performance superiority}. 
We compare SGS with other end-to-end TAD strategies in Tab.~\ref{tab:comparison_e2e}. From the aspect of memory usage, SGS leverages only 1.7 GB memory per video to train the model in an end-to-end fashion, which is much lower than other methods. From the aspect of detection performance, SGS reaches almost the same mAP as in naive end-to-end training, and beats other end-to-end strategies. For example, though TALLFormer~\cite{cheng2022tallformer} uses less forward computation by adopting the feature bank technique, the method may face the risk of failure if the training dataset is large, where the features of the same snippet between two epochs can be drastically different. Therefore, we insist to adopt the full forward propagation for all the snippets, and backward the gradient sequentially and selectively.

\vspace{4pt}\noindent {\bf{SGS is complementary to other memory-saving techniques.}}
We also compare SGS with other memory-saving techniques. For example, activation checkpointing \cite{chen2016training} also saves part of intermediate activations and does forward recomputation during backpropagation, but it operates on the model's different layers. Mixed-precision technique \cite{micikevicius2017mixed} adaptively combines half-precision computation  to save memory and speed up the training. The gradient accumulation sums the gradient over multiple batches to implicitly change the batch size to save memory. For comparison, our sequential gradient updating process focuses on reducing the complexity over temporal dimensions, instead of over batch dimensions or depth dimensions. And the proposed gradient sampling further reduces the backprop computation without sacrificing the action detection performance. Overall, SGS is designed for long-form end-to-end video understanding, and is complementary to the aforementioned three memory-saving techniques. It allows the detector to accommodate higher resolution frames, larger batch sizes, and/or a deeper video encoder. 

\begin{table}[t]
% \vspace{-0.5cm}
	\centering
	\caption{\textbf{Comparison of Sequentialized Gradient Sampling with other end-to-end training strategies in TAD.} We set the sampling rate to 30\% and use ETAD detector in all experiments. Computation in forward/backward stands for the theoretical computation cost of the video encoder during each propagation. GPU memory (per video) is reported with the TSM-ResNet50 backbone.}
% \vspace{-0.2cm}
	\small
 \setlength{\tabcolsep}{2pt}
	\begin{tabular}{p{3.2cm}p{1.1cm}<{\centering}p{1.3cm}<{\centering}p{0.85cm}<{\centering}>{\columncolor[gray]{0.9}}p{0.9cm}<{\centering}}
	\toprule
	Methods    & \makecell{Forward}    & \makecell{Backward}  & \makecell{Mem.} & \makecell{mAP} \\
	\midrule     
    Pre-extracted Feature        &  0\%       & 0\%      &  1.2GB  & 36.13 \\ 
	\midrule     
    Multi-stage Training \cite{xu2020g}  & 60\%   & 60\%  & 11GB & 36.36 \\
    Feature Bank \cite{cheng2022tallformer} & 30\%       & 30\%        & 11GB & 36.54 \\
    \textbf{SGS (ours)} & \textbf{130\%} & \textbf{30\%} & \textbf{1.7GB} & \textbf{36.79} \\
    \midrule
    Naive End-to-End & 100\% & 100\%  & 34GB &36.85 \\
    \bottomrule
	\end{tabular}
	\label{tab:comparison_e2e}
% \vspace{-0.3cm}
\end{table}

\section{Conclusion}
\label{sec:conclusion}
In this paper, we propose an end-to-end training method for the temporal action detector with extremely low GPU memory consumption. The training pipeline of ETAD contains sequentialized video encoding, action detector learning, and sequentialized gradient updating. 
ETAD achieves state-of-the-art action detection performance on multiple benchmarks. The proposed sequentialized gradient sampling method makes end-to-end training tractable in real-world applications, and the empirical results of different sampling strategies can shed light on how to effectively reduce computations in video localization problems. We hope this work will encourage the community to carry out more research on end-to-end training in various untrimmed video understanding tasks, such as video language grounding and video captioning.

%%%%%%%%% REFERENCES
{\small
\bibliographystyle{ieee_fullname}
\bibliography{egbib}
}

\renewcommand\thesection{\Alph{section}}
\renewcommand\thesubsection{\thesection.\arabic{subsection}}
\setcounter{section}{0}
\newpage
\section{Detailed Architecture of ETAD}
\label{supp:etad_arch}

In this section, we describe the detailed designs of two modules in ETAD: feature enhancement module and proposal evaluation module. Then, we introduce the loss function of our model, and more implementation details. 

\subsection{Feature Enhancement Module} 
\begin{figure}[ht]
\centering
% \vspace{-0.1cm}
\includegraphics[width=1\linewidth]{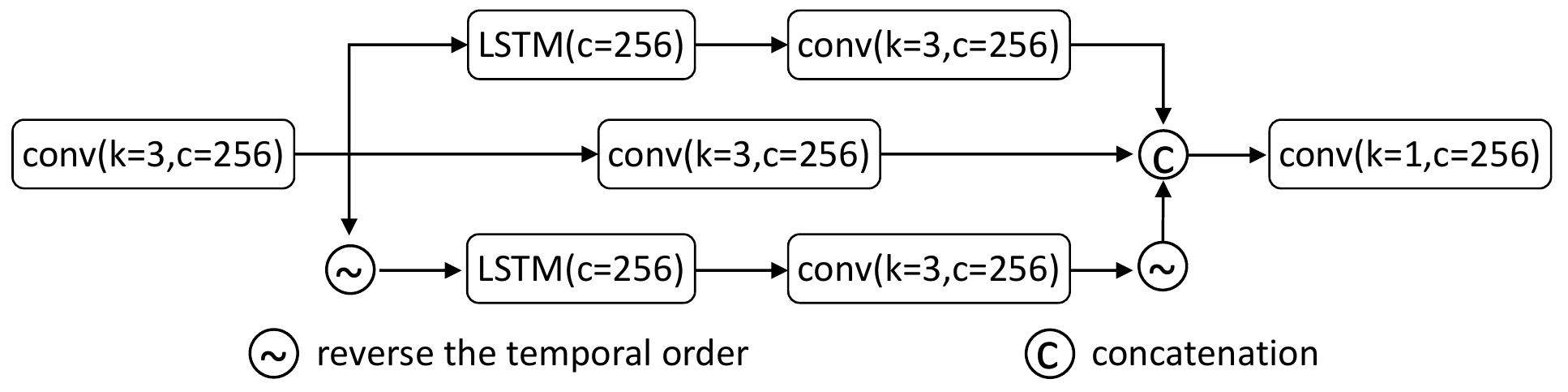}
% \vspace{-0.5cm}
\caption{\textbf{Architecture of feature enhancement module.} K, C stand for kernel size and channel number of corresponding layer.}
\label{fig:supp_feature_arch}
\end{figure}

As shown in Fig. \ref{fig:supp_feature_arch}, feature enhancement module adopts two LSTM \cite{hochreiter1997long} with different aggregation directions to capture both forward and backward context. The residual connection in the middle can mitigate the effect of forgetting issue brought by the LSTM. Group normalization layer with group number 16 and ReLU are used after each convolution layer. We also study its effectiveness by replacing it with a convolutional network (Conv) and a vanilla transformer. In Tab.~\ref{tab:ablation_feature_module} (top), Transformer shows the lowest mAP, since it requires more data to converge and stronger regularization to optimize. Conv also shows low mAP due to its inability to capture long-range context. Compared with the above two, our LSTM-based module shows the best performance.

\begin{table}[ht]
\centering
\caption{\textbf{Ablation of different feature enhancement modules (Feat. Layer), and different number of proposal evaluation modules (\#PEM)} on ActivityNet-1.3 with TSM-R50.}
% \vspace{-0.3cm}
\small
\begin{tabular}{p{1.7cm}p{0.8cm}<{\centering}|p{0.7cm}<{\centering}p{0.7cm}<{\centering}p{0.71cm}<{\centering}>{\columncolor[gray]{0.7}}p{1cm}<{\centering}}
\toprule
Feat. Layer & \#PEM & 0.5 & 0.75 & 0.9 &  avg. \\
\midrule
Transformer          & 1 & 52.75 & 36.34 & 7.28 & 35.34  \\ 
Conv   & 1 &  53.06 & 36.72 & 7.32 & 35.71 \\
LSTM          & 1 &  53.52 & 37.54 & 6.08 & \textbf{36.10} \\
\midrule
LSTM          & 2 & 54.02 & 37.84 &7.96   & 36.59  \\ 
LSTM          & 3 & 53.79 & 37.59 & 10.56 & \textbf{36.79} \\
LSTM          & 4 & 53.26 & 37.65 & 9.65 & 36.57  \\ 
\bottomrule
\label{tab:ablation_feature_module}
\end{tabular}
% \vspace{-1cm}
\end{table}

\subsection{Proposal Evaluation Module} 
The architecture of proposal evaluation module is shown in Tab. \ref{table:pem_structure}. Given a candidate proposal set, we first use the interpolation and rescaling algorithm in G-TAD \cite{xu2020g} as RoI alignment to extract the proposal features. Then we refine the proposals with several FC layers from three aspects: (1) The offset of the predicted start/end boundary. (2) The offset of the predicted center/width. (3) The IoU score between the proposal with the ground truth. What's more, we find adding a branch to classify the proposal startness and endness is helpful for IoU regression. After one proposal evaluation module, proposals will be refined by the average of start/end offset  and center/width offset.

To further improve the boundary precision of predicted actions, we follow the cascade-RCNN~\cite{cai2018cascade} to stack three proposal evaluation modules, where the proposals generated by the first stage are further refined in the second stage and so forth.
We use the increased IoU thresholds for the three stages, namely 0.7, 0.8, and 0.9. As such, the proposal boundaries are expected to become more accurate after each stage, which is also proved in  Tab. \ref{tab:ablation_feature_module}. 
It is worth mentioning that our cascade proposal refinement does not rely on an additional  proposal generation network, which is different from \cite{qing2021temporal}. 

\begin{table}[ht]
\centering
\caption{\textbf{The detailed architecture of proposal evaluation module.} $N$ is the number of candidate action proposals.}
\small
% \vspace{-0.2cm}
\begin{tabular}{p{1cm}<{\centering}|p{1cm}<{\centering}p{1.3cm}<{\centering}|p{2.1cm}}
\toprule
\multicolumn{4}{c}{\textbf{Proposal Start/End Offset Regression}} \\
\hline
layer &  dim & act & output size \\
\hline
\multicolumn{3}{c|}{proposal start/end feature} & $128 \times 8 \times N$  \\
\hline
$FC$   & 512 & $relu$  & $512 \times N$ \\
\hline
$FC$   & 128 & $relu$  & $128 \times N$ \\
\hline
$FC$   & 128 & $relu$  & $128 \times N$ \\
\hline
$FC$   & 128 & $relu$  & $128 \times N$ \\
\hline
$FC$   & 1   & $\times$ & $1 \times N$ \\
\hline
\end{tabular}
\begin{tabular}{p{1cm}<{\centering}|p{1cm}<{\centering}p{1.3cm}<{\centering}|p{2.1cm}}
\toprule
\multicolumn{4}{c}{\textbf{Proposal Center/Width Regression}} \\
\hline
layer &  dim & act & output size \\
\hline
\multicolumn{3}{c|}{proposal extended feature} & $128 \times 32 \times N$  \\
\hline
$FC$   & 512 & $relu$  & $512 \times N$ \\
\hline

$FC$   & 128 & $relu$  & $128 \times N$ \\
\hline
$FC$   & 128 & $relu$  & $128 \times N$ \\
\hline
$FC$   & 128 & $relu$  & $128 \times N$ \\
\hline
$FC$   & 2   & $\times$ & $2 \times N$ \\
\hline
\end{tabular}
\begin{tabular}{p{1cm}<{\centering}|p{1cm}<{\centering}p{1.3cm}<{\centering}|p{2.1cm}}
\toprule
\multicolumn{4}{c}{\textbf{Proposal IoU Regression}} \\
\hline
layer &  dim & act & output size \\
\hline
\multicolumn{3}{c|}{proposal extended feature} & $128 \times 32 \times N$  \\
\hline
$FC$   & 512 & $relu$  & $512 \times N$ \\
\hline
$FC$   & 128 & $relu$  & $128 \times N$ \\
\hline
$FC$   & 128 & $relu$  & $128 \times N$ \\
\hline
$FC$   & 128 & $relu$  & $128 \times N$ \\
\hline
$FC$   & 2   & $sigmoid$ & $2 \times N$ \\
\hline
\end{tabular}
\label{table:pem_structure}
% \vspace{-0.5cm}
\end{table}

\subsection{Loss function.}
The loss function of our method consists of boundary evaluation loss and cascade proposal refinement loss. $\mathcal{L}$ is computed as follows:
\begin{equation}
    \mathcal{L} =
    \mathcal{L}_{ce:bd_s} + 
     \sum_{i=1,2,3} \left(
    \mathcal{L}_{ce:bd_p}^i + \mathcal{L}_{iou}^i + \lambda\mathcal{L}_{rg:secw}^i \right)
\end{equation}
where $i$ is the index of the cascade proposal evaluation module, and weight $\lambda$ is set to 10 for balancing the losses.

$\mathcal{L}_{ce:bd_s}$ in the boundary evaluation module uses batch-level positive-negative-balanced cross entropy to supervise the startness or endness of each snippet, which is the same as proposed in BMN \cite{lin2019bmn}. We use the same loss for $\mathcal{L}_{ce:bd_p}$ in proposal evaluation module to compute the cross entropy of the proposal's startness and endness. Using $\mathcal{L}_{ce:bd_p}$ is helpful for stabilizing the learning of IoU confidence. $\mathcal{L}_{iou}$ contains a classification loss and a regression loss for the predicted IoU, which follows \cite{lin2019bmn}. The classification loss is cross-entropy loss, and the regression loss is L2 loss. For $\mathcal{L}_{rg:secw}$, we use the smooth-L1 loss for regressing start/end offset and center/width offset. We only do regression on positive samples, and the threshold of positive samples is gradually improved in the cascaded proposal evaluation module, \ie  0.7, 0.8, 0.9. 

\subsection{Implementation Details}
\label{supp:imple_details}

\noindent{\textbf{Training.}} In ActivityNet-1.3, we resize the feature sequences to a fixed length of 128 snippets. For THUMOS-14, we sample the features per 4 frames with fps 30, and utilize the sliding window approach with window length 128 and stride 64 for videos to generate training samples.  

\noindent{\textbf{Inference.}} To post-process network outputs, we use the boundary selecting method in \cite{lin2019bmn} to select proposals with high startness and endness, and use the averaged proposal boundary generated from three proposal evaluation modules. Soft-NMS is adopted based on proposal confidence scores $p=p_s \cdot p_e \cdot p_{iou}$, where $p_s$ and $p_e$ are from $\mathcal{L}_{ce:bd_s}$ standing for the start and end probabilities of a proposal, and $p_{iou}$ is the IoU score of the proposal from $\mathcal{L}_{iou}$. 

\section{Effectiveness of APS}
\subsection{APS in End-to-end training}
In Fig.4 of the main paper, we compared the performance with different APS ratios given pre-extracted features. Here, we also evaluate our method with different APS ratios under end-to-end training. As shown in Tab.~\ref{tab:aps_e2e}, end-to-end training generally improves the detection performance if the APS ratio is larger than 2\%, and the performance also starts to saturate with larger ratios. However, if the APS ratio is larger than 10\%, the mAP becomes lower. This is because too many proposals would cause the learning bias of the training dataset (\eg large proposals in ActivityNet). Thus, we choose 6\% as the APS ratio by default. 

\begin{table}[ht]
\setlength{\tabcolsep}{3pt}
\centering
% \vspace{-0.1cm}
\caption{\textbf{Ablations of different APS ratio with end-to-end training} on ActivityNet-1.3 with TSM-R50.}
\small
\label{tab:aps_e2e}
% \vspace{-0.3cm}
\begin{tabular}{c|cccccccc}
	\toprule
	E2E        & 0.1\% & 0.2\%  & 2\%  & 4\%  & \textbf{6\%} & 10\% & 20\% & 100\%\\
	\midrule
	\xmark      & 34.59 & 35.22 & 35.70 & 35.93 & \textbf{36.13} & 36.09 & 36.07 & 36.10  \\
	\cmark      & 34.50 & 35.21 & 36.34 & 36.41 & \textbf{36.79} & 36.72 & 36.66 & 36.51 \\
	\bottomrule
	\end{tabular}
% \vspace{-0.2cm}
\end{table}

\subsection{APS during inference}
As the default, ETAD only performs APS during training to reduce the computation cost. In inference, we use all the predicted proposals for higher detection performance. However, as a tool for selecting proposals, APS can also be applied during inference. Based on such motivation, we adopt grid sampling strategy with APS during inference, and Tab.~\ref{tab:infer} shows that APS is also effective for reducing inference complexity while preserving accuracy. Only the sampling ratio is smaller than 10\%, the mAP starts to decrease visibly. We did not conduct APS in inference as the default, considering that the impact of different APS ratios is rather small for the inference time and the inference GFLOPs in end-to-end setting.

\begin{table}[ht]
	\centering
% 	\vspace{-0.2cm}
	\caption{\textbf{Ablations of different APS ratios during inference} on ActivityNet-1.3 with TSM-R50.}
    % \vspace{-0.3cm}
	\small
	\label{tab:infer}
	\begin{tabular}{m{1.4cm}|m{0.9cm}<{\centering}m{0.7cm}<{\centering}m{0.7cm}<{\centering}m{0.7cm}<{\centering}m{0.7cm}<{\centering}}
		\toprule
		APS Ratio               & \textbf{100\%} & 20\%  & 15\%  & 10\%  & 6\% \\
		\midrule
		mAP                     & \textbf{36.79} & 36.77 & 36.74  & 36.71 & 36.51   \\
		\bottomrule
	\end{tabular}
% 	\vspace{-0.3cm}
\end{table}

\begin{table*}[ht]
\centering
\caption{\textbf{Comparison of ETAD with other state-of-the-art methods on HACS with same pre-extracted SlowFast features.} Total GPU memory with batch size 16 is reported.}
% \vspace{-0.3cm}
\small
\begin{tabular}{p{4cm}p{1cm}<{\centering}p{1cm}<{\centering}p{1cm}<{\centering}>{\columncolor[gray]{0.9}}p{2cm}<{\centering}p{2cm}<{\centering}p{2cm}<{\centering}}
	\toprule
	Methods           & 0.5   & 0.75  & 0.95  & Avg. mAP  &  Memory (GB) & Training Time \\
	\midrule     
    BMN~\cite{lin2019bmn}                & 52.49 & 36.38 & 10.37 & 35.76 & 12.10 & 58 min\\
    BMN~\cite{lin2019bmn} + TCANet~\cite{qing2021temporal}                & 55.60 & 40.01 & 11.47 & 38.71 & 12.34 & 104 min \\ 
    \midrule
    \textbf{ETAD} (random)      & 55.71 & 39.06 & 13.78 & \textbf{38.77}  & \textbf{3.28} & \textbf{50 min}\\
    ETAD (grid)        & 55.49 & 39.09 & 14.08 & 38.76  & 3.28 &  50 min\\ 
    ETAD (block)      &  51.46  & 34.26  & 11.43  & 34.49 & 3.28 &  50 min \\ 
    \bottomrule
\end{tabular}
\label{tab:hacs_results}
% \vspace{-0.3cm}
\end{table*}

\begin{table*}[ht]
\centering
\caption{\textbf{Comparison of different gradient sampling ratio $\gamma$ on THUMOS test set.} The GPU memory is reported by each video.}
% \vspace{-0.3cm}
\small
\begin{tabular}{p{3cm}p{1cm}<{\centering}p{0.7cm}<{\centering}p{0.7cm}<{\centering}p{0.7cm}<{\centering}p{0.7cm}<{\centering}>{\columncolor[gray]{0.9}}p{1.5cm}<{\centering}p{1.9cm}<{\centering}p{1.9cm}<{\centering}}
	\toprule
	Feature Encoder              & 0.3  & 0.4  & 0.5 & 0.6  & 0.7  & Avg. mAP  & Memory (GB) & Training Time  \\
	\midrule     
    SlowOnly ($\gamma\!=\!0\%$)   & 52.45 & 44.11 & 34.32 & 24.84 & 15.89 & 34.32 & - & -\\ 
    SlowOnly ($\gamma\!=\!10\%$)  & 59.72 & 52.74 & 42.73 & 32.98 & 23.02 & 42.23 & 1.06 & 2.08 h\\ 
    SlowOnly ($\gamma\!=\!30\%$)  & 60.66 & 52.87 & 42.95 & 33.31 & 23.38 & 42.63 & 1.06 & 2.25 h\\ 
    SlowOnly ($\gamma\!=\!100\%$) & 60.18 & 52.93 & 44.40 & 33.88 & 23.76 & 43.03 & 1.06 & 3.09 h\\ 

	\midrule     
    TSM ($\gamma\!=\!0\%$)        & 52.18 & 42.80 & 33.10 & 24.20 & 14.05 & 33.26 & - & - \\ 
    TSM ($\gamma\!=\!10\%$)       & 57.63 & 48.76 & 38.12 & 28.55 & 18.39 & 38.28 & 1.19 &  2.23 h\\ 
    TSM ($\gamma\!=\!30\%$)       & 56.50 & 49.16 & 39.17 & 29.47 & 19.07 & 38.67 & 1.19 & 2.51 h \\ 
    TSM ($\gamma\!=\!100\%$)      & 57.44 & 48.99 & 39.55 & 29.37 & 18.60 & 38.79 & 1.19 & 3.92 h \\ 
    \bottomrule
\end{tabular}
\label{tab:thumos_e2e}
% \vspace{-0.5cm}
\end{table*}

\section{Results on HACS dataset}
\label{supp:hacs_result}
We also report the results of ETAD on HACS \cite{zhao2019hacs} dataset based on the pre-extracted SlowFast feature, since this is a fair comparison with other state-of-the-art methods that use the same feature. HACS is a recent large-scale temporal action localization dataset, containing 140K action instances from 50K videos including 200 action categories. In  this dataset, we adopt SlowFast \cite{slowfast_iccv19} features provided by \cite{qing2021temporal} and rescale the feature sequences to 224 snippets. The only training difference from ActivityNet is that we use the learning rate of $4\times 10^{-4}$  and batch size of 16 on HACS.

As shown in Tab.~\ref{tab:hacs_results}, ETAD can outperform the baseline method BMN~\cite{lin2019bmn} by a large margin. Compared with state-of-the-art method TCANet \cite{qing2021temporal}, ETAD can also achieve comparable performance. (1) Particularly, the training time is visibly reduced from 104 mins to 50 mins, and the GPU memory decreases from 12.34 GB to 3.28 GB. This further proves the existence of proposal redundancy in TAD and the effectiveness of our APS design. (2) ETAD  also exceeds TCANet on the high IoU threshold scenario by 2.5\%, which is similar to ActivityNet-1.3 and THUMOS14. (3) What's more, TCANet relays on the proposal generation result from BMN, while our single model does not need any extra proposal generation network, suggesting the simplicity of ETAD.
(4) At last, we also test different proposal sampling strategies on HACS and find the results are similar to those in ActivityNet. Both random sampling and grid sampling achieve decent performance. Since the block sampling breaks the distribution of different proposals, thus the detection performance is much worse than others. 

% \vspace{-0.1cm}
\section{More results on THUMOS dataset}
\label{supp:e2e_thumos}

To further verify the effectiveness of proposed gradient sampling in SGS, we also test different gradient sampling ratios on THUMOS dataset under end-to-end training, as shown in Tab.~\ref{tab:thumos_e2e}. Here, we choose SlowOnly-R50 or TSM-R50 as the feature encoder. In this ablation, since we are using shorter clip (8 frames per clip), lower resolution (180$\times$180), and only RGB modality, the performance is expected to be lower than the state-of-the-art performance listed in Tab.2 of the main paper.

From the results, we can find that if the gradient sampling ratio $\gamma$ is 0, \ie frozen encoder, the performance is not that promising. However, once we unfreeze the backbone, the performances are instantly boosted with more than 7\% gains of average mAP using SlowOnly, and more than 5\% gains of average mAP using TSM. This verifies the importance of end-to-end training again. Besides, with different sampling ratios, we interestingly find the performances generally remain at the same level. Such a conclusion is consistent with the ablation results on ActivityNet-1.3 dataset, further proving that a small portion of snippets is enough for end-to-end training in TAD and the effectiveness of our gradient sampling approach.

\section{Additional Study of End-To-End Training}
\label{supp:e2e_study}
In this section, we discuss several factors that are important for the video encoder in end-to-end training, such as data augmentation, frozen backbones, frame resolution, and pretraining. For ablation, we remove the sequential design for all experiments in this section to reflect the real GPU memory consumption.

\noindent\textbf{Data augmentation is vital for end-to-end training.} One of the main advantages of end-to-end training is that we can use data augmentation on original frames, which is not possible in feature-based settings. As shown in Tab. \ref{tab:e2e_more_frozen} (top), we implement random cropping and temporal jittering at the snippet level as data augmentation. Here, temporal jittering means we shift a random stride of each frame in a snippet. Compared with not using data augmentation, random cropping is very helpful for TAD while temporal jittering slightly harms the performance. Therefore, we only use random cropping in our experiment as default.

\noindent\textbf{Partially freeing backbone can have a good trade-off between computation and performance.} It is a common trick to save the computation by freezing some shallow layers of video encoder. In this study, we want to know how  the frozen layers affect the detection performance. For a ResNet-based encoder (\eg TSM) with four stages, we can gradually freeze the layers from shallow to deep. Tab.~\ref{tab:e2e_more_frozen} (middle) clearly shows that: \textbf{1)} End-to-end training is important for TAD, since the frozen stage of 4, \ie freeze the whole backbone, has the lowest mAP compared with others. \textbf{2)} As we freeze fewer encoder layers, detection performance will be improved, but the gain becomes smaller, and  the memory consumption also becomes much larger. \textbf{3)} To have a good trade-off between memory and performance, frozen stage 2 is recommended in our experiments.

\begin{table}[t]
	\centering
	\caption{\textbf{Study of different data augmentation, frozen stage, frame resolution, and pretraining of video encoder in end-to-end training} on ActivityNet-1.3. Note that SGS is not adopted in the experiments. We report the GPU memory usage per video with TSM-R50. $\dag$ means out of memory on a V100 GPU. $\ddag$ means the encoder is finetuned on ActivityNet by the classification task.}
	\small
% 	\vspace{-0.3cm}
	 \setlength{\tabcolsep}{2pt}
	\begin{tabular}{p{1.3cm}<{\centering}p{1.4cm}<{\centering}p{1.5cm}<{\centering}p{1cm}<{\centering}>{\columncolor[gray]{0.9}}p{1.1cm}<{\centering}p{1.1cm}<{\centering}}
	\toprule
	Encoder    & \makecell{Frame \\ Resolution} &  \makecell{Data \\ Augment.} & \makecell{Frozen\\ Stage} & \makecell{Average \\ mAP} & \makecell{Memory \\ (GB)}\\
	\midrule     
    TSM         & 112x112          &  $\times$          & 2      & 35.12 & 9.1 \\ 
    TSM         & 112x112          &  jitter            & 2      & 35.17 & 9.1 \\ 
    TSM         & 112x112          &  crop              & 2      & \textbf{35.53} & 9.1 \\ 
    TSM         & 112x112          &  crop+jitter       & 2      & 35.38 & 9.1 \\ 
	\midrule     
    TSM         & 112x112          &  crop              & 4      & 34.26 & 4.5 \\ 
    TSM         & 112x112          &  crop              & 3      & 35.01 & 4.6 \\ 
    TSM         & 112x112          &  crop              & 2      & \textbf{35.53} & 9.1 \\ 
    TSM         & 112x112          &  crop              & 1      & 35.52 & 17.0 \\ 
    TSM         & 112x112          &  crop              & 0      & 35.46 & 25.8 \\ 
    \midrule
    TSM         & 224x224          &  crop              & 4      & 36.24 & 17.5     \\ 
    TSM         & 224x224          &  crop              & 3      & 36.47 & 17.6 \\ 
    TSM         & 224x224          &  crop              & 2      & \textbf{36.79} & 34.3 \\ 
    TSM         & 224x224          &  crop              & 1      & - & OOM$\dag$   \\ 
    \midrule
    TSM-FT$\ddag$      & 224x224          &  crop              & 2      & \textbf{36.92} & 34.3 \\ 
    \bottomrule
	\end{tabular}
% 	\vspace{-0.5cm}
	\label{tab:e2e_more_frozen}
\end{table}

\noindent\textbf{Higher frame resolution can boost the performance by a large margin.} We also study the impact of the frame resolution on the detection performance in end-to-end training, as shown in Tab.~\ref{tab:e2e_more_frozen} (middle). If we freeze the whole backbone, a higher resolution can boost the mAP from 34.26 to 36.24 (+1.98), which is a significant improvement. If we freeze fewer encoder layers, the gap of mAP between low frame resolution and high frame resolution would be smaller, but still can bring +1.26 gains under frozen stage 2. However, the memory usage is almost doubled in this case. 

\noindent\textbf{Classification pretraining is not necessary if using end-to-end training.} Some other TAD methods  are proposed to finetune the video encoder by the classification task on the target dataset, then extract features~\cite{qing2021temporal}. In our case, if we finetune the encoder on ActivityNet by action recognition task, the detection performance would be slightly improved from 36.79 to 36.92, as shown in Tab.~\ref{tab:e2e_more_frozen} (bottom).
Such a small gain (+0.13) is reasonable, since end-to-end training can already improve the feature representation from the target domain. Since classification pretraining also takes a long time to train, thus it is not 
necessary to conduct this pretraining when end-to-end training is adopted.

To summarize, we recommend  giving priority to  higher frame resolution with stronger data augmentation when with a limited resource budget in end-to-end training. If necessary, we can further freeze certain layers in the encoder to save computation. We believe such a study for end-to-end training will enlighten the TAD community in the sense of efficiency and efficacy trade-off.

\begin{algorithm}[t]
\caption{PyTorch-like Pseudocode of Sequentialized Gradient Sampling. }
\definecolor{codeblue}{rgb}{0.25,0.5,0.5}
\lstset{
  backgroundcolor=\color{white},
  basicstyle=\ttfamily\footnotesize,
  columns=fullflexible,
  breaklines=true,
  captionpos=b,
  commentstyle=\fontsize{8pt}{8pt}\color{codeblue},
  keywordstyle=\fontsize{8pt}{8pt},
}
\begin{lstlisting}[language=python]
# frames: Nx3xTxHxW, K: micro_batch size
optimizer.zero_grad()

# stage 1: sequentialized video encoding
feats = []
micro_batches = torch.chunk(frames, N//K, dim=0)
for micro_batch in micro_batches:
   with torch.set_grad_enabled(False):
      feat = self.video_encoder(micro_batch)
      feats.append(feat.detach())
feats = torch.stack(feats, dim=0)

# stage 2: action detector learning
feats.requires_grad(True).retain_grad()
with torch.set_grad_enabled(True):
   pred = self.action_detector(feats)
   loss = loss_func(pred, gt)
   loss.backward()
feats_grad = copy.deepcopy(feats.grad.detach())
    
# stage 3: sequentialized gradient sampling
sample_idx = torch.randperm(N)[:sample_size]
micro_batches = torch.chunk(frames[sample_idx], sample_size//K, dim=0)
grads = torch.chunk(feats_grad[sample_idx], sample_size//K, dim=0)
for micro_batch, grad in micro_batches, grads:
   with torch.set_grad_enabled(True):
      feat = self.video_encoder(micro_batch)
      feat.backward(gradient=grad)

# update the parameter
optimizer.step()
\end{lstlisting}
\label{alg:sgs}
\end{algorithm}

\section{Implementation details of SGS}
\label{supp:egs_implem}

In this section, we describe the implementation details of Sequential Gradient Sampling (SGS). As shown in Alg.~\ref{alg:sgs}, SGS can be divided into three stages. First, in sequentialized video encoding, the video is chunked (in temporal dimension) into multiple micro-batches, and feature of each micro-batch is sequentially extracted by the video encoder. 
Then, the action detector takes the concatenated features, and further, the parameters are updated by the loss. We collect the feature gradients and free all the cache in GPU
memory. Last, we sample a portion of feature gradients to save the computation (\eg random sampling), and backward the video encoder by a micro-batch. After sequentially backward all the sampled micro-batches, we accumulate all the gradients and update the video encoders' parameter.

\section{Details of Feature-guided Sampling}
\label{supp:fps_dps}

In the paper, we explore two feature-guided sampling strategies, \ie \textit{farthest point sampling}, and \textit{determinantal point process}. The motivation of feature-guided sampling is that each data has different features than others, which means their inherent importance is also different from others. Therefore, one can try to leverage the information inside the data and ensure the most informative or important samples are selected. For TAD, we can adopt feature-guided sampling on the embedding space of samples (\eg snippet features, proposal features). The pseudocode can be found in Alg.~\ref{alg:fps_dpp}.

The farthest point sampling (FPS) is a common feature-guided sampling approach, which has been adopted in many fields such as point cloud understanding. Given the data points $X\!\in\! {{\mathbb{R}}^{N \times C}}$, where $N$ is the number of total samples, and $C$ is the dimension number of each sample feature, FPS selects the new point from the unselected points, and ensures new point has the farthest distance to the currently selected data points in the embedding space. The distance between two different points can be measured by a distance function, which we use euclidean distance in our case. Such sampling actually samples the next point in the middle of the least-known area of the sampling domain, and thus can guarantee the sampled points are most distinguished from each other. However, since this sampling process is conducted iteratively, thus the corresponding time complexity is $O(N^2)$.

\begin{algorithm}[t]
\caption{Pseudocode of FPS/DPP sampling strategy. }
\label{alg:fps_dpp}
\definecolor{codeblue}{rgb}{0.25,0.5,0.5}
\lstset{
  backgroundcolor=\color{white},
  basicstyle=\ttfamily\footnotesize,
  columns=fullflexible,
  breaklines=true,
  captionpos=b,
  commentstyle=\fontsize{8pt}{8pt}\color{codeblue},
  keywordstyle=\fontsize{8pt}{8pt},
}
\begin{lstlisting}[language=python]
# data: N x C, sampling_ratio: (float) 0~1.
# sampling_strategy: fps or dpp.
import torch_cluster
from dppy.finite_dpps import FiniteDPP

N = data.shape[0]
# farthest point sampling
if sampling_strategy == "fps":
   index = torch_cluster.fps(data, ratio=sampling_ratio)

# determinantal point process
if sampling_strategy == "dpp":
   data = np.float64(data)
   sample_num = int(sampling_ratio * N)
   # likelihood kernel, use eye matrix to increase the rank
   kernel = data.dot(data.T) + 1e-2 * np.eye(N) 
   DPP = FiniteDPP("likelihood", **{"L": kernel})
   index = DPP.sample_exact_k_dpp(size=sample_num)
    
# return the index list of selected samples
\end{lstlisting}
\end{algorithm}

We also implement another feature-guided sampling as the determinantal point process (DPP). DPP measures the sample probability as a determinant of some kernels. In our case, we use cosine similarity as the kernel function, and update the likelihood matrix every iteration. Since our sampling ratio is fixed, we can use kDPP \cite{kulesza2011k} to approximate DPP for fast sampling. To meet the requirements of kDPP, we add an eye matrix filled with small values (\eg 1e-2) to ensure the rank of the likelihood matrix is larger than the sample size. In general, DPP sampling improves the diversity of sampled data in the embedding space.

\begin{figure}[ht]
    \centering
	\includegraphics[trim={1.2cm 0cm 1.9cm 0cm},width=1.0\linewidth]{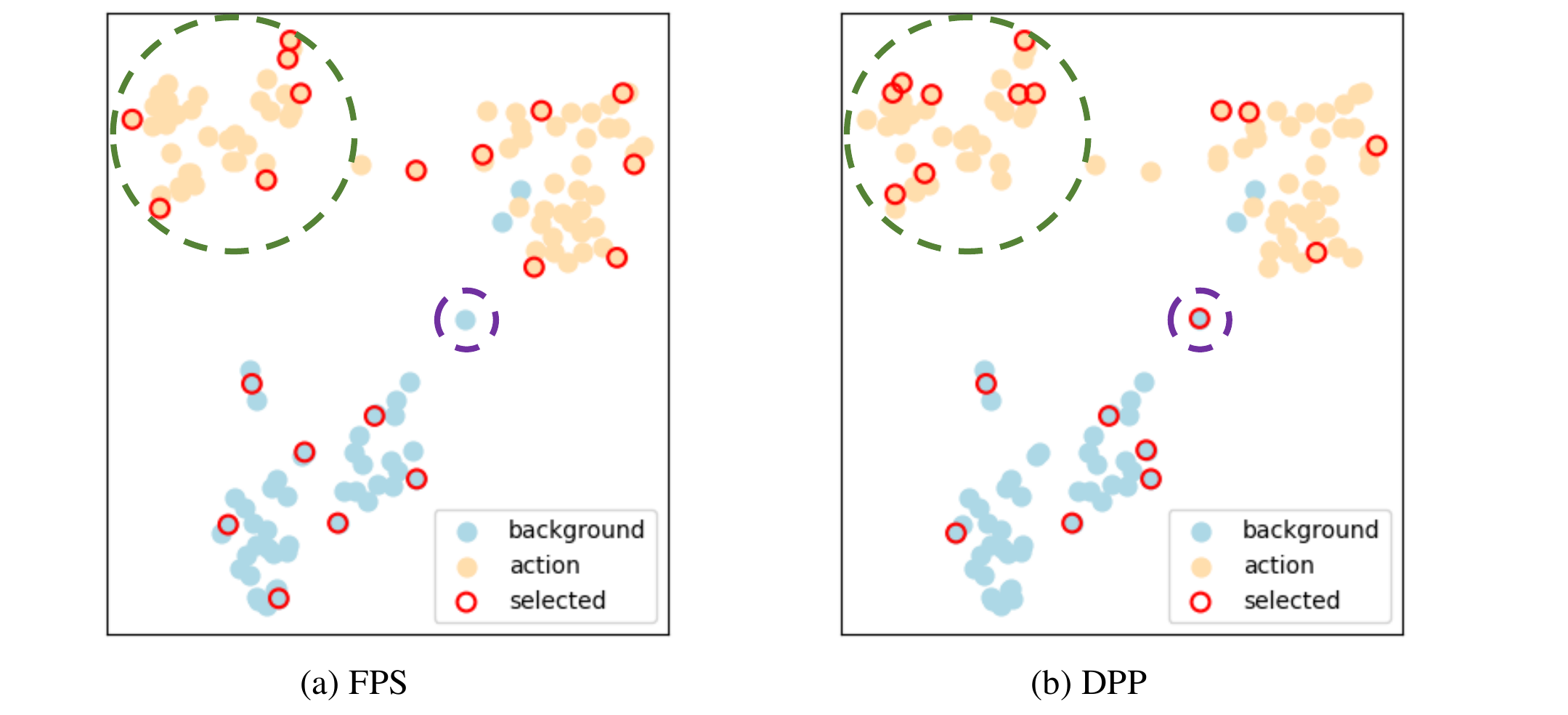}
    \caption{\textbf{t-SNE visualization of FPS sampling and DPP sampling}. The orange dots are snippets inside action and the blue dots are background snippets. Dots with red outlines are the sampled snippets.}
    \label{fig:supp_fps_dpp}
% \vspace{-0.5cm}
\end{figure}

In our experiments, we find that DPP works always better than FPS, as shown in Tab.4 in the main paper. To further analyze these two strategies, we provide the t-SNE visualization \cite{van2008visualizing} of FPS and DPP at the snippet level, as shown in \Figref{fig:supp_fps_dpp}. In this figure, the yellow points are the action foreground snippets, and the blue points are the background snippets. 
Initially, we can find that these snippets can be well grouped in different clusters based on their features, which verifies the necessity of conducting sampling on feature embedding space. We observe that \textbf{(1)} For points in the dashed green circle, FPS tends to select extreme points, while DPP can select samples with a larger variety. \textbf{(2)} For the point in the purple dash circle, FPS misses such a hard-negative sample because its distance from other points is not that big in the embedding space. However, such points may be informative samples, and DPP successfully selects this representative sample. Those two findings can explain the success of DPP for snippet-level gradient sampling.

For DPP and FPS on the proposal level, we notice that FPS would prefer to focus on  small-scale proposals since these proposal features are more distinguished from each other in the feature space. Due to a lack of enough learning of middle and large-scale proposals, FPS behaves badly in proposal sampling. On the contrary, DPP can sample all scale proposals and succeed in this case.
\begin{figure*}[t]
% \vspace{-0.6cm}
    \centering
	\includegraphics[width=0.9\linewidth]{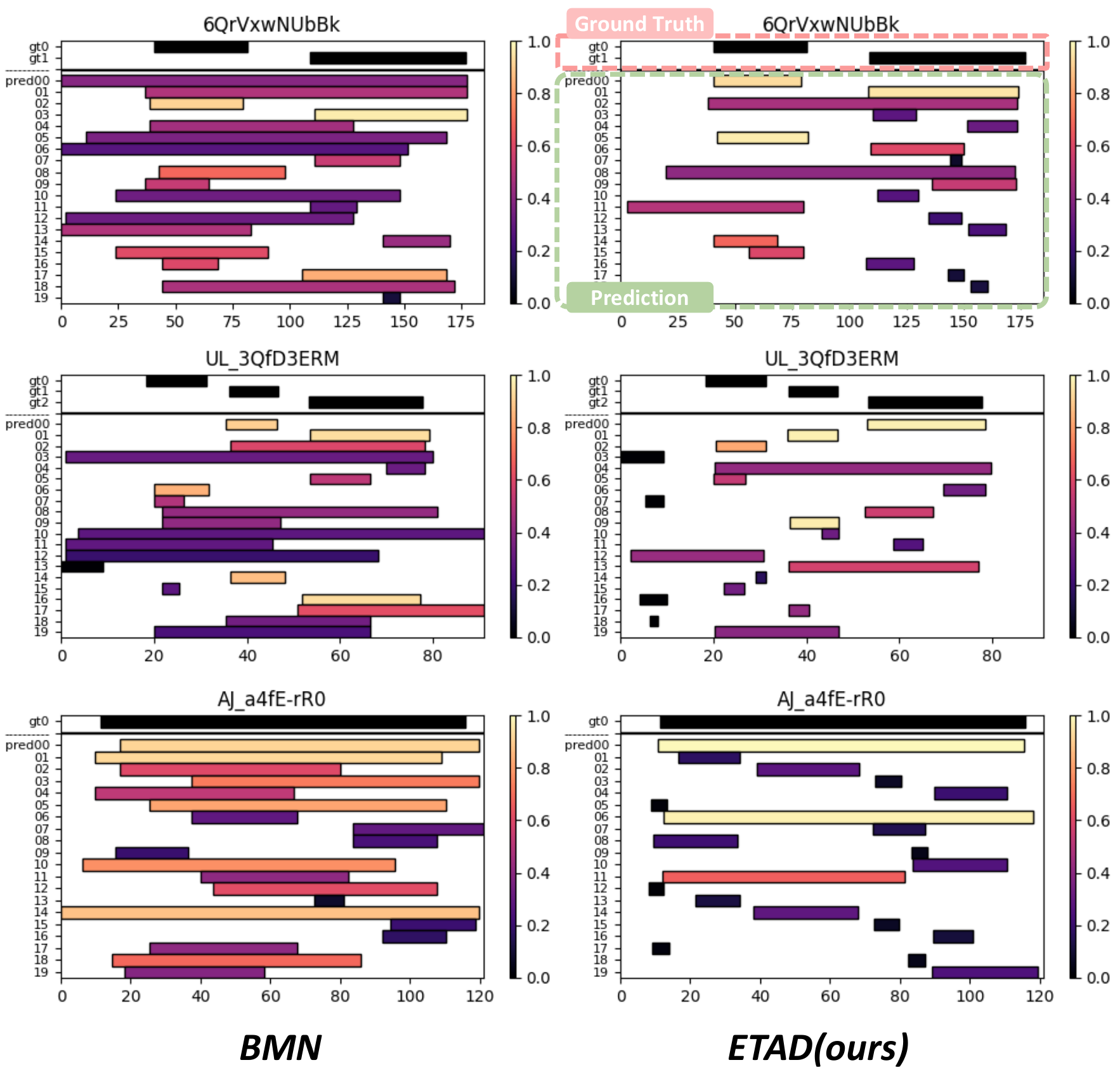}
    % \vspace{-0.5cm}
    \caption{Qualitative results of ETAD and BMN on ActivityNet-1.3. The color of the proposal represents the maximum IoU of this proposal to ground truth actions. We plot the ground truth actions of each video (drawn in black and above the black line), and top-20 predicted proposals by algorithms (drawn in colors and under the black line).}
    \label{fig:quatitive_result}
    % \vspace{-0.3cm}
\end{figure*}

\section{Qualitative Visualization}
\label{supp:visual}

In order to provide a more vivid understanding of our method, we visualize the qualitative predictions of our method and BMN \cite{lin2019bmn} on ActivityNet for comparison. In \Figref{fig:quatitive_result}, we plot the ground truth actions of each video (drawn in black and above the black line), and also the top-20 predicted proposals by algorithms (drawn in colors and under the black line). The color of the proposal represents the maximum IoU of this proposal to the ground truth actions. Therefore, a proposal with lighter color means it has more overlap with the ground truth, indicating this is a high-quality proposal.

As demonstrated in the figure, ETAD can generate \textbf{(1) more precise proposal boundary.} For instance, in the first and third row in \Figref{fig:quatitive_result}, the boundary of proposals from ETAD is closer to the real action boundary than BMN. \textbf{(2) more reliable proposal confidence.} As shown in the first and second row in \Figref{fig:quatitive_result}, ETAD has fewer false positive proposals and proves that regressed proposal confidence is much more reliable than BMN, indicating the advantage of our method on proposal ranking.

\end{document}